\pdfoutput=1

\documentclass[11pt]{article}

\usepackage[final]{acl}

\usepackage{times}
\usepackage{latexsym}
\usepackage{enumitem}
\usepackage{wrapfig}
\usepackage{amsmath}  
\usepackage{colortbl} 
\usepackage{xcolor}   
\usepackage{array}
\usepackage{booktabs}
\usepackage{subcaption}
\usepackage{url}     
\usepackage{xurl}   
\usepackage{graphicx}
\usepackage{adjustbox}
\definecolor{pastelblue}{RGB}{222, 235, 247}
\definecolor{pastelgreen}{RGB}{223, 240, 216}
\definecolor{pastelpink}{RGB}{252, 228, 236}
\definecolor{headergray}{RGB}{200, 200, 200}

\usepackage[T1]{fontenc}

\usepackage[utf8]{inputenc}

\usepackage{microtype}

\usepackage{inconsolata}

\usepackage{graphicx}

\usepackage{hyperref}   
\usepackage{fontawesome5} 
\usepackage{graphicx}   
\usepackage{scalerel}   
\usepackage{xspace}     
\usepackage{natbib}
\usepackage{acl}

\title{T-VEC: A Telecom-Specific Vectorization Model with Enhanced Semantic Understanding via Deep Triplet Loss Fine-Tuning}

\newcommand{\huggingface}{\scalerel*{\includegraphics[height=1em]{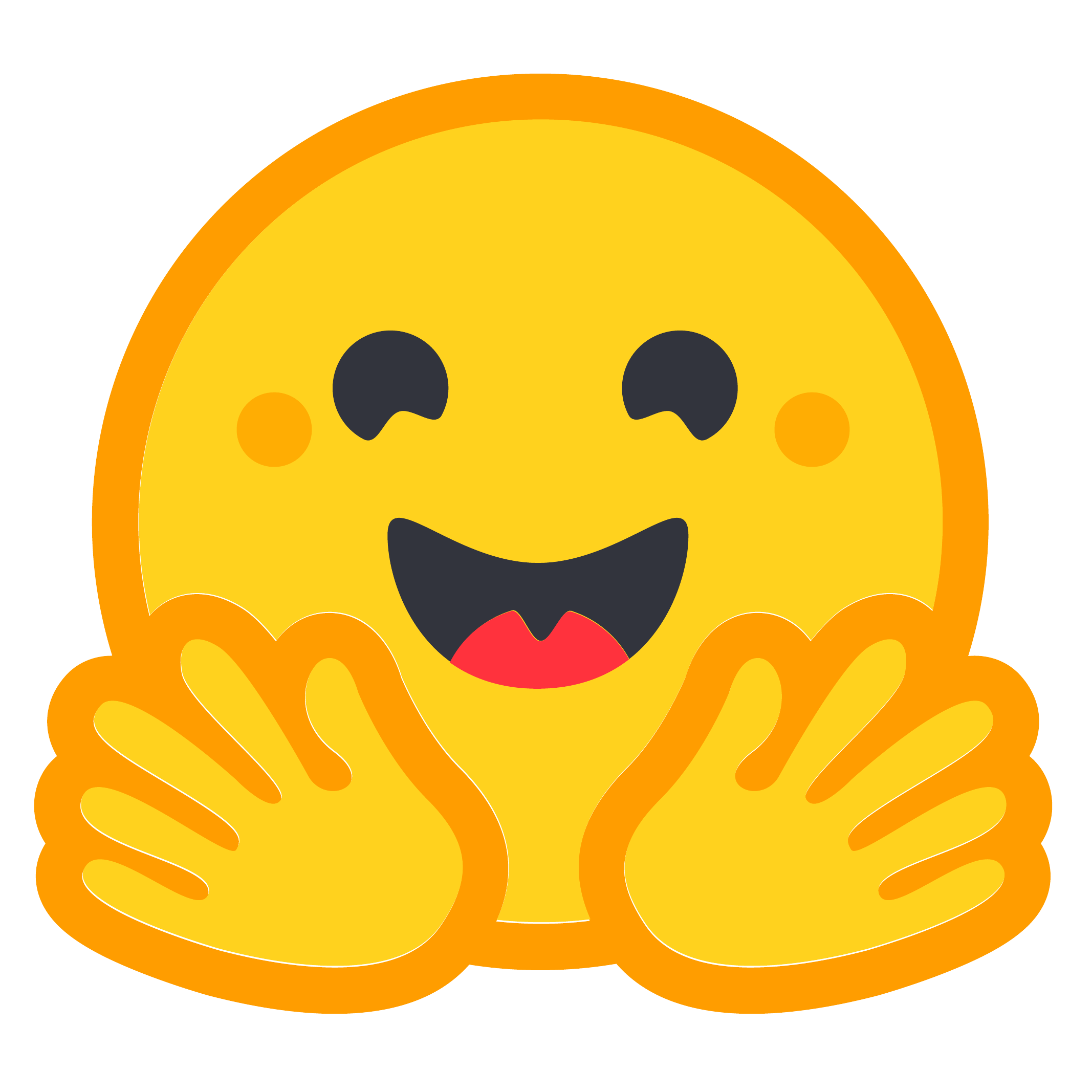}}{\textrm{C}}\xspace}

\author{
\begin{tabular}{ccc}
Vignesh Ethiraj\thanks{Equal contribution} & Ashwath David\footnotemark[1] & Sidhanth Menon\footnotemark[1] \\
\multicolumn{3}{c}{Divya Vijay\footnotemark[1] \quad Vidhyakshaya Kannan\footnotemark[1]} \\
\end{tabular} \\[1em] 
\texttt{\{vignesh.e, ashwath.d, sidhanth.m, divya.v, vidhyakshaya.k\}@netoai.ai} \\[0.3em] 
NetoAI \\[0.5em] 
\huggingface~\href{https://huggingface.co/NetoAISolutions/T-VEC/tree/main}{Model} \quad
\huggingface~\href{https://huggingface.co/datasets/NetoAISolutions/TEmbed}{Data}
}

\begin{document}
\maketitle
\begin{abstract}
The specialized vocabulary and nuanced concepts of the telecommunications industry pose persistent challenges for standard Natural Language Processing (NLP) models. Generic embedding models often struggle to represent telecom-specific semantics, limiting their utility in retrieval and downstream tasks. We present \textsc{T-VEC} (Telecom Vectorization Model), a domain-adapted embedding model fine-tuned from the \texttt{gte-Qwen2-1.5B-instruct} backbone using a triplet loss objective. Fine-tuning was performed on T-Embed, a high-quality, large-scale dataset covering diverse telecom concepts, standards, and operational scenarios. Although T-Embed contains some proprietary material and cannot be fully released, we open source 75\% of the dataset to support continued research in domain-specific representation learning. On a custom benchmark comprising 1500 query-passage pairs from IETF RFCs and vendor manuals, \textsc{T-VEC} surpasses MPNet, BGE, Jina and E5, demonstrating superior domain grounding and semantic precision in telecom-specific retrieval. Embedding visualizations further showcase tight clustering of telecom-relevant concepts. We release \textsc{T-VEC} and its tokenizer to support semantically faithful NLP applications within the telecom domain. 
\end{abstract}

\section{Introduction}

Text embeddings—dense vector representations of text—serve as the backbone for many modern NLP applications, including semantic search, dialogue systems, and information retrieval \cite{reimers-gurevych-2019-sentence}. While general-purpose models such as BERT \cite{devlin-etal-2019-bert} and GPT-2 \cite{radford2019language} have shown strong performance on broad benchmarks, their effectiveness often degrades in specialized technical domains characterized by domain-specific jargon, overloaded terminology, and structural ambiguity \cite{gururangan-etal-2020-dont}.

Telecommunications exemplifies such a domain. It features an unusually dense mix of acronyms (e.g., \textit{MME}, \textit{SMF}, \textit{gNB}), technical jargon (\textit{handover}, \textit{QoS parameters}), and ambiguous terms (\textit{cell}, \textit{sector}, \textit{core}), many of which carry very different meanings in general contexts. This linguistic complexity is further amplified by rapidly evolving standards (e.g., \textit{5G}, \textit{LTE}, \textit{NFV}) and layered architectures (e.g., \textit{RAN}, \textit{core}, and \textit{transport networks}).

Despite its real-world importance, telecommunications remains underserved in NLP research. Existing models struggle to accurately interpret telecom language, limiting performance in tasks like fault log analysis, technical document retrieval, customer intent classification, and regulatory compliance. Addressing this domain-language gap is vital for deploying effective AI solutions in operational telecom environments.

To bridge this gap, we introduce \textbf{T-VEC (Telecom Vectorization Model)}, a domain-adapted sentence embedding model trained via deep triplet loss fine-tuning. Our core contributions are threefold:

\begin{figure*}[t]
\centering
\includegraphics[width=\textwidth]{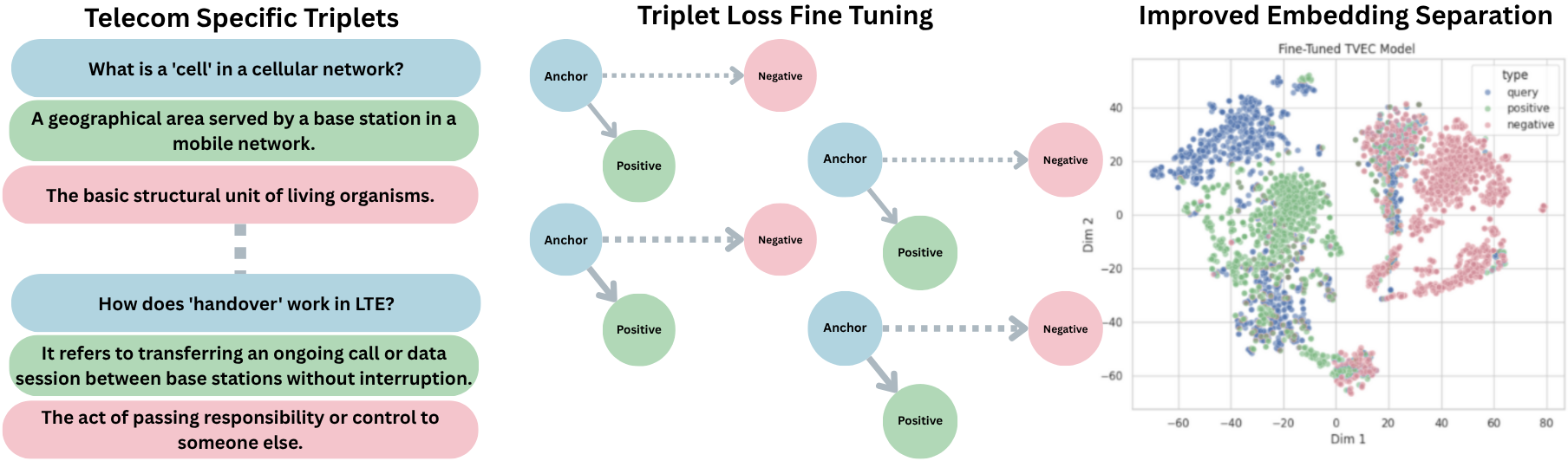} 
\vspace{0.5em}
\caption{\textbf{From noisy telecom jargon to meaningful machine understanding.} T-VEC learns telecom semantics by training on curated triplets: a domain-specific query (anchor), a true paraphrase (positive), and a deceptive distractor (negative). Through triplet loss fine-tuning, the model learns to pull related meanings closer while pushing apart unrelated ones, resulting in clear, telecom-aware clusters in embedding space.}
\label{fig:tvec-full-width}
\end{figure*}

\begin{enumerate}

\item \textbf{T-Embed.}
We construct a high-quality, large-scale telecom dataset, T-Embed, covering diverse telecom concepts, standards, and operational contexts. Although the dataset contains proprietary information and the full dataset cannot be publicly released, we open source 75\% of it (MIT license). 

\item \textbf{Open-Source Domain-Specific Embedding Model.}  
We release \textsc{T-VEC}, the first open-source embedding model specialized for the telecommunications domain. \textsc{T-VEC} is obtained via full-model fine-tuning of \texttt{gte-Qwen2-1.5B-instruct} using a triplet loss objective, with updates across all \textbf{338 transformer layers}. This yields domain-aligned representations for retrieval and semantic matching tasks in telecom. The trained model is publicly available to support reproducibility and real-world applications.

\item \textbf{Open-Source Telecom Tokenizer.}  
We release the first tokenizer tailored to telecom corpora and integrate it into \textsc{T-VEC}. Built by extending the \texttt{gte-Qwen2-1.5B-instruct} tokenizer with domain-specific vocabulary, it improves token segmentation and subword coverage for telecom acronyms, protocol names, and configuration terms. Shared tokens retain their original embeddings, while new tokens are randomly initialized and updated during fine-tuning. This approach enables \textsc{T-VEC} to represent telecom jargon more accurately without distrupting the pretrained model.

\end{enumerate}

Our comprehensive evaluations demonstrate that T-VEC achieves state-of-the-art performance on standard benchmarks (leading MTEB average score) while exhibiting superior understanding of telecom semantics compared to its base model and other strong general-purpose models on our domain-specific benchmarks.

\section{Related Work}

Generating effective text representations is a fundamental challenge in NLP. Sentence-Transformers \cite{reimers-gurevych-2019-sentence} popularized the use of siamese network structures with pre-trained models like BERT \cite{devlin-etal-2019-bert} to create semantically meaningful sentence embeddings. Subsequent research has produced numerous powerful general-purpose embedding models, including MPNet-based models (\texttt{all-mpnet-base-v2}) \cite{NEURIPS2020_c3a690be}, E5 (\texttt{e5-base-v2}) \cite{wang2024multilingual}, BGE (\texttt{bge-base-en-v1.5}) \cite{chen2024bge}, GTE (\texttt{gte-base}, now including Qwen2-based variants like our base model) \cite{li2023towards}, Jina Embeddings \cite{jina2023embeddings2}, and instruction-tuned models like Instructor \cite{su-etal-2023-one}. These models excel on general language tasks due to training on vast, diverse web corpora.

However, the limitations of general models in specialized domains are well-documented \cite{gururangan-etal-2020-dont, tang2025finmteb}. While domain-specific embeddings have been extensively studied in healthcare \cite{alsentzer-etal-2019-publicly, 10.1093/bioinformatics/btz682}, finance \cite{anderson-etal-2024-greenback}, accelerator physics \cite{hellert2024physbert} engineering \cite{braun2021language}, cybersecurity \cite{roy2017learning} and law \cite{chalkidis-etal-2020-legal}, there has been little progress in creating or evaluating telecom-specific text embeddings. Previous work \cite{roychowdhury2024telecom} includes a detailed study of domain-adapted sentence embeddings in the telecom sector, emphasizing the challenges and methods for effective document retrieval. Despite this work, public telecom datasets and standardized benchmarks comparable to those in other domains remain scarce. Standard evaluation suites such as MTEB \cite{muennighoff2022mteb} do not include telecom standards, network logs, or regulatory filings, and telecom-oriented benchmarks are largely absent. As a result, research and development of domain-adapted embedding models for telecommunications has lagged behind other high-impact verticals, motivating our release of T-VEC and its supporting telecom-specific evaluation resources. 

Fine-tuning using objectives like triplet loss \cite{Schroff_2015_CVPR} is particularly effective for learning fine-grained semantic similarity relevant to tasks like search and retrieval within a specific domain. While some domain adaptations might only involve fine-tuning the final layers or adding small adapter modules, our work pursues deep fine-tuning, modifying a significant portion of the base model's weights to fundamentally reshape its representational space for the target domain.

\section{Methodology}

\subsection{Base Model}

Our base model is \texttt{gte-Qwen2-1.5B-instruct} from Alibaba-NLP\footnote{\url{https://huggingface.co/Alibaba-NLP/gte-Qwen2-1.5B-instruct}}, a 1.5B-parameter transformer producing 1536-dimensional embeddings. It supports sequences up to 32K tokens and belongs to the Qwen2 family \cite{bai2023qwen}, known for strong multilingual and instruction-following capabilities. We fine-tune this model to specialize its representations for telecom-specific tasks.

\subsection{Curating T-Embed: A Telecom Triplet Embedding Dataset}

The fine-tuning dataset, T-Embed, was curated by a team of experienced telecommunications professionals. This process was designed to ensure that the resulting dataset would not only be large in scale, but also exhibit the depth, breadth, and nuance required to capture the full complexity of telecom language, operations, and standards. 

The final dataset comprises \textbf{100,000 triplets}, capturing a wide spectrum of telecom knowledge, including thousands of unique concepts, procedures, and system artifacts. To support open research in domain-adapted representation learning, we publicly release \textbf{75,000 triplets} (75\% of the dataset).

Further details on T-Embed, including token statistics, topic-wise query distribution, and vocabulary characteristics, are provided in Appendix~\ref{sec:fine_tuning_dataset_details}.

\subsubsection{Topic and Subdomain Coverage}
The curation process began with an exhaustive mapping of the telecommunications knowledge landscape. Experts systematically identified and catalogued all major and minor subdomains relevant to the industry, including but not limited to:

\begin{itemize}[noitemsep, topsep=0pt]
    \item \textbf{Wireless Technologies.} 3G, 4G/LTE, 5G NR, Wi-Fi, NB-IoT, and legacy standards.
    \item \textbf{Network Domains.} Radio Access Network (RAN), Core Network (EPC, 5GC), Transport (IP/MPLS, optical), Access (FTTx, DSL), and OSS/BSS.
    \item \textbf{Network Functions.} Coverage of both traditional (e.g., HSS, MME, SGW, PGW) and next-generation (e.g., AMF, SMF, UPF, AUSF, NRF, gNB, eNB) network elements.
    \item \textbf{Operational Procedures.} Fault management, alarm correlation, performance monitoring (KPI/KQI), configuration management, software upgrades, and network slicing.
    \item \textbf{Technical Documentation.} Vendor-specific manuals, 3GPP technical specifications, RFCs, ITU-T recommendations, and regulatory filings.
    \item \textbf{Emerging Topics.} O-RAN, virtualization (NFV, SDN), edge computing, private networks, and AI/ML for telecom.
\end{itemize}

This comprehensive taxonomy guided the balanced sampling of source materials, ensuring that both foundational and cutting-edge topics were sufficiently represented.

\subsubsection{Domain Vocabulary and Semantic Ambiguity}
Telecom language is characterized by dense layers of acronyms, abbreviations, and polysemous terms. The curation team placed special emphasis on vocabulary diversity and contextual disambiguation. For each subdomain, domain experts compiled extensive lists of:

\begin{itemize}[noitemsep, topsep=0pt]
    \item \textbf{Acronyms and Abbreviations.} e.g., ``MME'' (Mobility Management Entity), ``SMF'' (Session Management Function), ``gNB'' (next-gen NodeB), ``O-RAN'' (Open RAN).
    \item \textbf{Jargon and Technical Terms.} e.g., ``handover,'' ``RRC state,'' ``bearer,'' ``QoS parameter,'' ``paging,'' ``cell reselection,'' ``sector,'' ``slice.''
    \item \textbf{Ambiguous Terms.} Words with multiple meanings in telecom and general English (e.g., ``cell,'' ``core,'' ``sector,'' ``handover'').
\end{itemize}

Triplet construction explicitly targeted these terms to ensure the model would learn to resolve ambiguity based on context.

\subsubsection{Triplet Generation Methodology}
Let \(\mathcal{D} = \{(a_i, p_i, n_i)\}_{i=1}^N\) denote our curated corpus of triplets, where each triplet is defined as follows:
\[
(a, p, n) \;\in\; \mathcal{A} \times \mathcal{P} \times \mathcal{N},
\]
with
\begin{description}[noitemsep, topsep=0pt]
  \item[\(\displaystyle a\in\mathcal{A}\) (Anchor):]  
    A telecom‐specific input (e.g., a query, log entry, or protocol message) sampled from our domain corpus.
  \item[\(\displaystyle p\in\mathcal{P}\) (Positive):]  
    A text unit that is \emph{semantically equivalent} or \emph{contextually aligned} with \(a\), drawn from the same technical subdomain.
  \item[\(\displaystyle n\in\mathcal{N}\) (Negative):]  
    A text unit that is \emph{lexically or topically plausible} relative to \(a\) but \emph{semantically incorrect}, \emph{irrelevant}, or \emph{subtly misleading}.
\end{description}

The triplet generation process was iterative and multi-layered:

\begin{enumerate}[noitemsep, topsep=0pt]
    \item \textbf{Seed Collection.} Anchors were curated using a diverse corpus, including technical manuals, standards, incident tickets, and regulatory documents, as reference.
    \item \textbf{Positive Selection.} For each anchor, positives were manually paraphrased or retrieved using expert knowledge to ensure semantic closeness, often reflecting real-world telecom paraphrase phenomena (e.g., different vendor terminology for the same concept).
    \item \textbf{Negative Mining.} Negatives were not chosen at random; instead, ``hard negatives'' were prioritized. These are texts that are lexically or topically similar to the anchor but diverge in subtle, domain-relevant ways (e.g., confusing ``handover'' with ``cell reselection,'' or ``core'' with ``RAN'').
    \item \textbf{Quality Assurance.} Each triplet underwent review by at least two domain experts. Disagreements were resolved through discussion or further research. Ambiguous or low-quality triplets were iteratively refined or discarded.
\end{enumerate}

\subsection{Fine-Tuning with Triplet Loss}
We fine-tune the \texttt{gte-Qwen2-1.5B-instruct} model using the triplet loss objective \cite{Schroff_2015_CVPR} to encourage semantically meaningful embeddings. Given a triplet \((a,p,n)\in\mathcal{A}\times\mathcal{P}\times\mathcal{N}\), the model minimizes the following loss: 
\begin{multline}
L(a, p, n) = \max\big(0,\, d(E(a), E(p)) \\
\quad -\, d(E(a), E(n)) + \alpha \big)
\end{multline}
where $E(\cdot)$ denotes the embedding function, and $d(\cdot, \cdot)$ is the cosine distance, defined as
\[
d(x, y) = 1 - \cos(\theta_{x,y}),
\]
with $\cos(\theta_{x,y})$ being the cosine similarity between $x$ and $y$, and $\alpha$ is a margin hyperparameter. The objective ensures that the distance between anchor and negative exceeds that of the anchor and positive by at least $\alpha$. We construct triplets with query-like anchors (e.g., user questions or descriptions) to bias the model toward high retrieval performance on telecom-specific tasks.

\subsubsection{Deep Model Architecture Modifications}
A key differentiator of our approach lies in the depth of fine-tuning. Rather than constraining updates to lightweight adapters or a small subset of final layers, we perform end-to-end fine-tuning across \textbf{338 layers} of the \texttt{gte-Qwen2-1.5B-instruct} architecture. This enables substantial adaptation of internal representations, effectively reconfiguring a large fraction of the model's parameters to align with domain-specific semantics.

\subsubsection{Magnitude and Distribution of Weight Adaptation}  
To quantify the depth of fine‐tuning, we compute for each updated parameter tensor \(W\) the L2 norm  
\[
\Delta(W) \;=\;\bigl\lVert W_{\mathrm{fine}} - W_{\mathrm{base}}\bigr\rVert_{2}.
\]  
Across all modified tensors, the mean L2 change is  
\[
\overline{\Delta} \;=\; \frac{1}{M}\sum_{m=1}^{M}\Delta(W_{m}) \;=\; 0.7735,
\]  
indicating substantial redistribution of model capacity toward telecom‐specific features.  

Figure~\ref{fig:weight_change} visualizes the top-20 tensors with the largest \(\Delta(W)\). These tensors, spanning MLP gate, up‑ and down‑projection weights across layers 0–8, exhibit a broadly distributed adaptation pattern. The pervasiveness of these changes confirms that T‑VEC’s domain specialization arises from deep, architecture‑wide weight modifications rather than superficial surface tuning.

\section{Evaluation}

We evaluate T-VEC’s domain specialization primarily within the telecommunications domain. Our evaluation framework comprises three components: (1) a held-out test set of telecom triplets, (2) a domain-specific retrieval benchmark, and (3) embedding space analysis via similarity distributions and t-SNE projections. For completeness, we report T-VEC's performance on standard embedding benchmarks (e.g., STS, classification, MTEB tasks) in Appendix~\ref{sec:evaluation_on_standard_benchmarks}.

\subsection{Telecom Triplet Evaluation}
\begin{figure*}[t]
\centering
\includegraphics[width=0.48\linewidth]{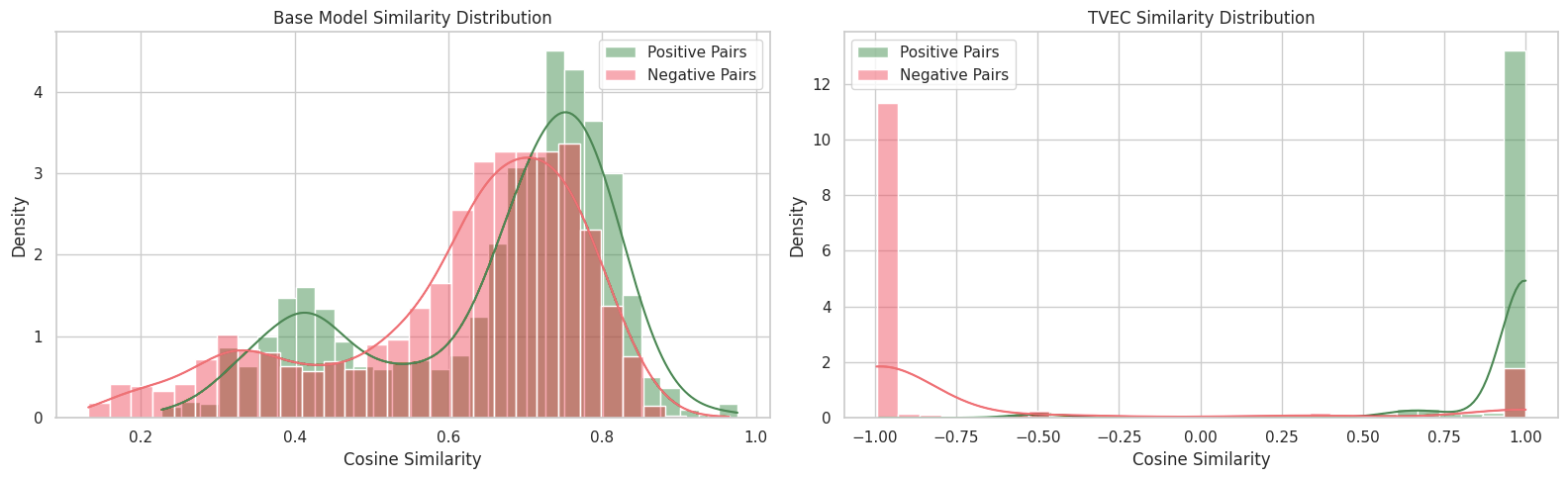}
\hfill
\includegraphics[width=0.48\linewidth]{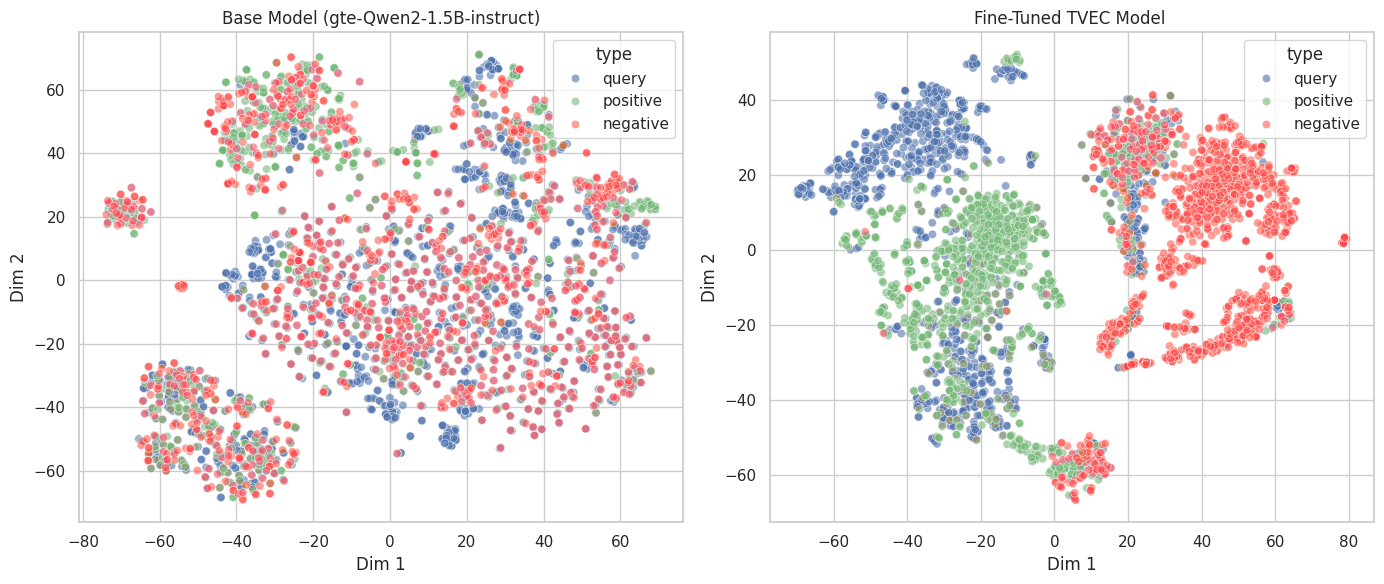}
\caption{
\textbf{Embedding space analysis.}
Left: Cosine similarity distributions for positive (green) and negative (red) telecom pairs. T-VEC (right) demonstrates clearer separation than the base model.
Right: t-SNE visualization of embeddings. T-VEC embeddings form tighter clusters with improved separation between anchor, positive, and negative samples.
}
\label{fig:embedding_analysis}
\end{figure*}

To assess the model’s ability to capture fine-grained semantic distinctions, we construct a held-out set of telecom triplets $(a, p, n)$ with no overlap with the training distribution. Each triplet consists of an anchor $a$, a semantically related positive $p$, and a plausible but incorrect negative $n$. The model is evaluated based on its ability to satisfy the triplet constraint:
\begin{equation}
d(E(a), E(p)) < d(E(a), E(n))
\end{equation}

where \( E(\cdot) \) is the embedding function and \( d(\cdot, \cdot) \) denotes cosine distance. Triplet accuracy is defined as the proportion of test triplets for which the constraint holds.

\begin{table}[h]
\centering
\caption{\textbf{Telecom-Specific Triplet Evaluation.} Accuracy on a held-out test set of telecom triplets measuring semantic discrimination. Each model is evaluated on its ability to embed the anchor closer to the positive than the negative in cosine space.}
\label{tab:telecom_triplet}
\begin{tabular}{l c}
\toprule
\textbf{Model} & \textbf{Triplet Accuracy} \\
\midrule
\textbf{T-VEC} & \textbf{0.9380} \\
GTE-Qwen2-1.5B-instruct & 0.0135 \\
all-mpnet-base-v2 & 0.0685 \\
bge-base-en-v1.5 & 0.0414 \\
e5-base-v2 & 0.0168 \\
jina-embeddings-v2-base-en & 0.0290 \\
instructor-xl & 0.0321 \\
gte-base & 0.0169 \\
multilingual-e5-base & 0.0120 \\
all-MiniLM-L6-v2 & 0.0637 \\
\bottomrule
\end{tabular}
\end{table}

T-VEC achieves a triplet accuracy of 0.9380, substantially outperforming both its base model and leading general-purpose embedding models. This indicates a robust ability to disambiguate nuanced telecom semantics.

\subsection{Telecom Retrieval Evaluation}
\label{sec:telecom_retrieval_evaluation}

\begin{table*}[t]
  \centering
  \caption{Comparison of cosine similarity-based evaluation metrics across embedding models.}
  \label{tab:cosine_model_comparison}
  \setlength{\tabcolsep}{4pt}
  \renewcommand{\arraystretch}{1.1}
 \begin{tabular}{lcccccccccc}
\toprule
\textbf{Metric} & \textbf{T-VEC} & \textbf{Qwen2} & \textbf{MPNet} & \textbf{BGE} & \textbf{E5} & \textbf{Jina} & \textbf{Instr.} & \textbf{GTE} & \textbf{mE5} & \textbf{MiniLM} \\
\midrule
CosineSim@1         & \textbf{0.78} & 0.72 & 0.70 & 0.69 & 0.69 & 0.67 & 0.71 & 0.68 & 0.69 & 0.65 \\
Avg\_CosineSim@5    & \textbf{0.74} & 0.70 & 0.67 & 0.66 & 0.67 & 0.64 & 0.68 & 0.63 & 0.65 & 0.61 \\
Top1\_CosineMatch   & \textbf{0.80} & 0.75 & 0.72 & 0.71 & 0.72 & 0.69 & 0.74 & 0.70 & 0.72 & 0.67 \\
Recall@5\_cosine    & \textbf{0.83} & 0.76 & 0.74 & 0.73 & 0.74 & 0.70 & 0.76 & 0.71 & 0.72 & 0.68 \\
\bottomrule
\end{tabular}

\end{table*}

To assess T-VEC’s effectiveness in domain-specific retrieval, we constructed a custom benchmark consisting of 1500 query-passage pairs derived from telecommunications documentation. The benchmark corpus comprises IETF RFCs\footnote{\url{https://www.rfc-editor.org/}} and vendor technical manuals\footnote{Scraped from publicly available documentation hosted on official vendor sites such as Cisco, Juniper, and Huawei.}. These documents were chosen for their authoritative status and technical specificity, making them ideal for evaluating retrieval systems that require precise semantic understanding in specialized domains.

RFCs (Requests for Comments) are public-domain specifications that define key protocols, architectures, and operational guidelines for the Internet and telecom infrastructure. They offer a rich source of structured, formal, and jargon-heavy content, which presents a meaningful challenge for semantic retrieval models. Similarly, vendor manuals often describe configuration parameters, troubleshooting workflows, and protocol extensions.
\paragraph{Preprocessing.}
We removed artifacts (e.g., ASCII drawings, null characters), deduplicated near-identical passages, and filtered out non-informative boilerplate content to ensure semantic quality and relevance.
This benchmark provides a realistic and challenging testbed for evaluating retrieval in specialized technical domains, where understanding precise semantics and domain terminology is essential.
Each document was segmented into semantically coherent chunks using structural markers such as section headers and paragraph boundaries. These chunks served as the unit of retrieval. To simulate realistic information needs, we used a large language model (LLM) to generate one query per chunk. Each chunk was passed as input to the LLM, which returned a corresponding query that is topically and semantically aligned with the content of the chunk. The exact prompting strategy is detailed in Appendix~\ref{sec:fine_tuning_dataset_details}.
Each query is paired with its originating (ground-truth) passage, along with several hard negatives sampled from the same corpus to encourage fine-grained semantic discrimination.
We evaluated a range of publicly available embedding models on this benchmark using cosine similarity-based retrieval. As shown in Table~\ref{tab:cosine_model_comparison}, T-VEC outperformed all other models across all metrics, including CosineSim@1, Recall@5, and top-1 match rate.

\subsection{Embedding Space Analysis}

To analyze the semantic geometry of T-VEC’s learned embedding space, we visualize cosine similarity distributions and t-SNE plots (Figure~\ref{fig:embedding_analysis}).Positives are tightly grouped near high cosine similarity values, while negatives remain well separated. This confirms that the model has effectively internalized domain-specific semantics.

\section{Real-World Deployment}
Our domain-specific embedding model has been integrated into a production-grade platform that supports chat-based interaction with a growing corpus of internal and external documents. It enables users to retrieve precise, context-aware answers from a collection of organizational knowledge bases, forum discussions, and technical documentation.

The chatbot uses our custom-trained embedding model to improve retrieval performance for specialized terminology and nuanced queries that general-purpose models often struggle with. The model powers dense retrieval over a hybrid index (dense + sparse), ensuring high recall and semantic fidelity. It has been optimized for performance in noisy, real-world environments with domain-specific jargon, abbreviations, and informal user queries.

The chatbot has indexed over 10,000 documents across various formats (e.g., PDFs, Markdown), and supports multi-document reasoning via chunked embedding aggregation. The chatbot interface has handled over 50,000 queries in pilot deployments, with human evaluation suggesting significant improvement in answer relevance over baseline models such as \texttt{text-embedding-ada-002}.

 User insights are being fed back into a continuous retraining loop, allowing the embedding model and retrieval logic to co-evolve with real user interactions.

Deployment challenges included latency optimization, query disambiguation, and integrating user feedback into the model improvement cycle. We addressed these via efficient vector search infrastructure (FAISS with GPU support), prompt-tuning pipelines, and lightweight feedback interfaces embedded into the chat UI.

Overall, this deployment demonstrates the viability and impact of domain-adapted embeddings in augmenting enterprise productivity.

\section{Conclusion}
We introduced T-VEC, a 1.5B parameter telecom-specific text embedding model derived from \texttt{gte-Qwen2-1.5B-instruct}. Through extensive and deep fine-tuning on a large, manually curated telecom dataset using triplet loss, T-VEC achieves state-of-the-art performance on telecom-specific semantic understanding tasks, significantly outperforming general models and its own base model.

We release T-VEC and its telecom-specific tokenizer under the MIT license to support transparency, reproducibility, and broader adoption across telecom and NLP communities. 

Future work includes expanding the T-VEC fine-tuning dataset with even more diverse telecom data, exploring architectural enhancements specifically for telecom NLP tasks, and deploying T-VEC in real-world telecom applications to further validate and refine its capabilities.

\section*{Limitations}

While T-VEC demonstrates strong performance on telecom-specific tasks, it exhibits notable limitations in terms of generalization to broader natural language understanding. As shown in Table~\ref{tab:allnli}, T-VEC underperforms significantly on the \texttt{sentence-transformers/all-nli} triplet benchmark, achieving an average triplet score of only 0.6150—far below general-purpose sentence embedding models like \texttt{all-mpnet-base-v2} or \texttt{bge-base-en-v1.5}. This degradation reflects a well-documented trade-off in domain-adaptive fine-tuning~\cite{gururangan-etal-2020-dont, howard-ruder-2018-universal, lee2020biobert}: models optimized for in-domain semantic distinctions often lose representational flexibility when applied to out-of-distribution (OOD) contexts.

This over-specialization is likely driven by T-VEC’s intensive fine-tuning on telecom triplets, which may have narrowed its semantic space to focus exclusively on patterns, terminology, and structure relevant to telecommunications. While this narrowing enables precise disambiguation and ranking within the target domain, it reduces the model’s ability to encode more abstract, general-purpose semantic relationships that are crucial in tasks like natural language inference, paraphrase detection, and commonsense reasoning.

Additionally, our fine-tuning did not employ strategies such as multi-domain pretraining, continual learning, or regularization techniques (e.g., feature drift control or contrastive mixing) to explicitly preserve generalization. Exploring such methods is a promising direction for future work, especially for applications that demand both high in-domain precision and robust zero-shot generalization. 

While T-VEC demonstrates strong retrieval gains, retrieval evaluation was conducted on a domain-specific benchmark of limited size (1500 query passage pairs). The test queries were LLM-generated from telecom texts; broader evaluation on human-authored queries and larger retrieval pools remains future work. Quantitative analysis in real-world deployment scenarios, at scale, or on human-authored queries has not been performed. Broader evaluation is a key avenue for future work.

Finally, there are practical integration constraints. \textsc{T-VEC}’s embedding dimensionality must match the downstream system (e.g., 768 dimensions for many Transformer-based chatbots). Mismatched dimensions require projection layers or model retraining, introducing latency and deployment complexity. Another promising direction is to explore smaller, more efficient models for deployment, which could reduce memory footprint and inference time while retaining strong in-domain performance. Implementing parameter-efficient approaches such as Low-Rank Adaptation (LoRA) could enable more scalable experimentation while reducing overfitting risks. Exploring such methods is a promising direction for future work, especially for applications that demand both high in-domain precision and robust zero-shot generalization.


\bibliography{custom}

\appendix
\begin{table*}[htbp]
\caption{Example triplets targeting telecom-specific ambiguity and jargon. Each triplet illustrates disambiguation of acronyms, polysemous terms, or technical vocabulary.}
\label{tab:ambiguous_triplets}
\begin{tabular}{@{}p{0.2\textwidth}p{0.37\textwidth}p{0.37\textwidth}@{}}
\toprule
\rowcolor{white}
\textbf{Query} & \textbf{Positive Response} & \textbf{Negative Response} \\
\midrule
\rowcolor{pastelblue}
What is the function of the SMF in 5G core networks? & 
\cellcolor{pastelgreen}It handles session management and IP address allocation for user equipment. & 
\cellcolor{pastelpink}It measures signal strength and adjusts beam direction in the RAN. \\
\rowcolor{white}
How is handover managed during inter-gNB transitions? & 
The RRC handles signaling for seamless user mobility between gNBs. & 
It encodes the user’s location in the IP header for routing. \\
\rowcolor{pastelblue}
What is a bearer in the context of LTE? & 
\cellcolor{pastelgreen}A bearer is a virtual channel with specific QoS parameters assigned to data flows. & 
\cellcolor{pastelpink}A bearer is the physical antenna that transmits radio signals. \\
\rowcolor{white}
How does paging work in idle RRC state? & 
The network sends paging messages to wake idle UEs when there's incoming data. & 
Paging refers to dynamic spectrum sharing between different frequency bands. \\
\rowcolor{pastelblue}
What role does the core play in 5G slicing? & 
\cellcolor{pastelgreen}The core ensures logical isolation and resource allocation across slices. & 
\cellcolor{pastelpink}The core handles beamforming and antenna configuration at the edge. \\
\bottomrule
\end{tabular}
\end{table*}

\section{Fine-Tuning Dataset Details}
\label{sec:fine_tuning_dataset_details}
We present summary statistics of our fine-tuning dataset in Table~\ref{tab:token_stats}, reporting token-level metrics across query, positive, and negative components. This includes average, minimum, and maximum token counts, as well as the overall vocabulary size computed across all fields. In Table~\ref{tab:ambiguous_triplets}, we show example triplets selected to target telecom-specific challenges, including domain-specific acronyms, polysemous terminology (e.g., “core,” “cell”), and industry jargon. These triplets are curated to explicitly test the model’s ability to resolve ambiguity in context.
\begin{figure}[htbp]
    \centering
    \includegraphics[width=0.85\linewidth]{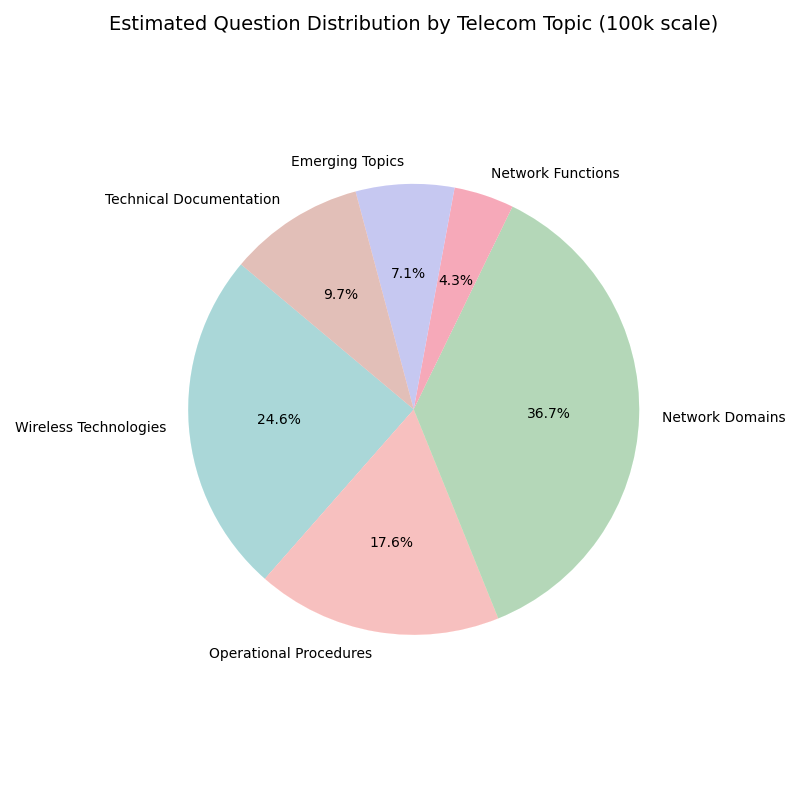}
    \caption{Estimated distribution of telecom-related queries across key annotated topic categories.}
    \label{fig:topic_distribution}
\end{figure}

\begin{figure*}[htbp]
    \centering
    %
    \begin{subfigure}[b]{0.49\textwidth}
        \centering
        \includegraphics[width=\linewidth]{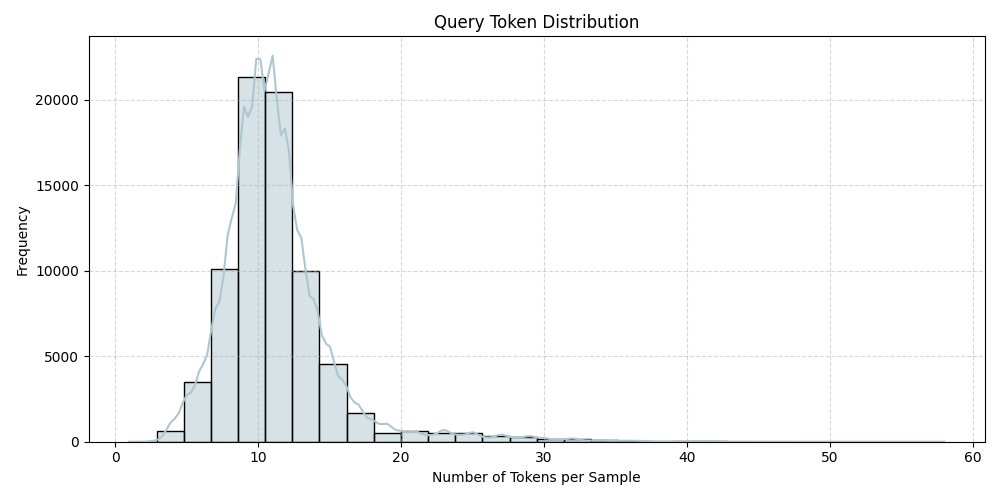}
        \caption{Query tokens}
        \label{fig:query_token_dist}
    \end{subfigure}
    \hfill
    %
    \begin{subfigure}[b]{0.49\textwidth}
        \centering
        \includegraphics[width=\linewidth]{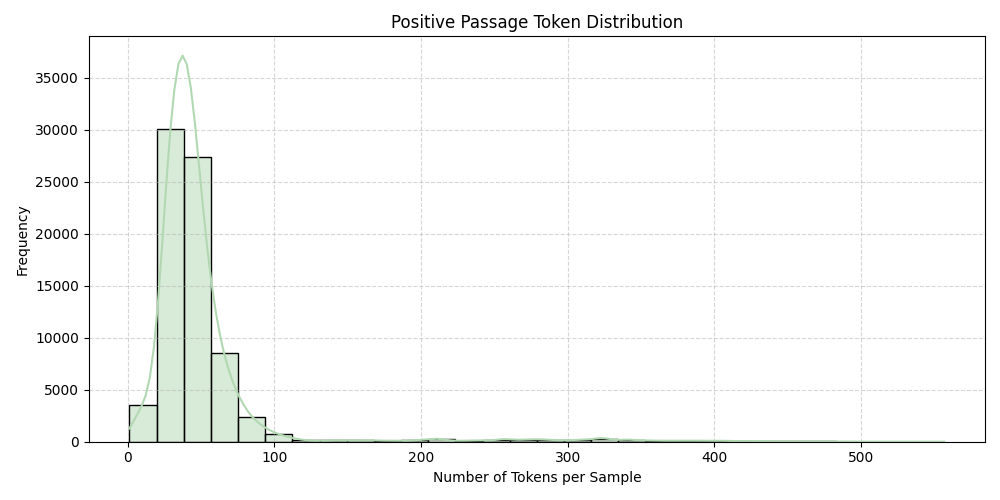}
        \caption{Positive tokens}
        \label{fig:pos_token_dist}
    \end{subfigure}
    \hfill
    %
    \begin{subfigure}[b]{0.49\textwidth}
        \centering
        \includegraphics[width=\linewidth]{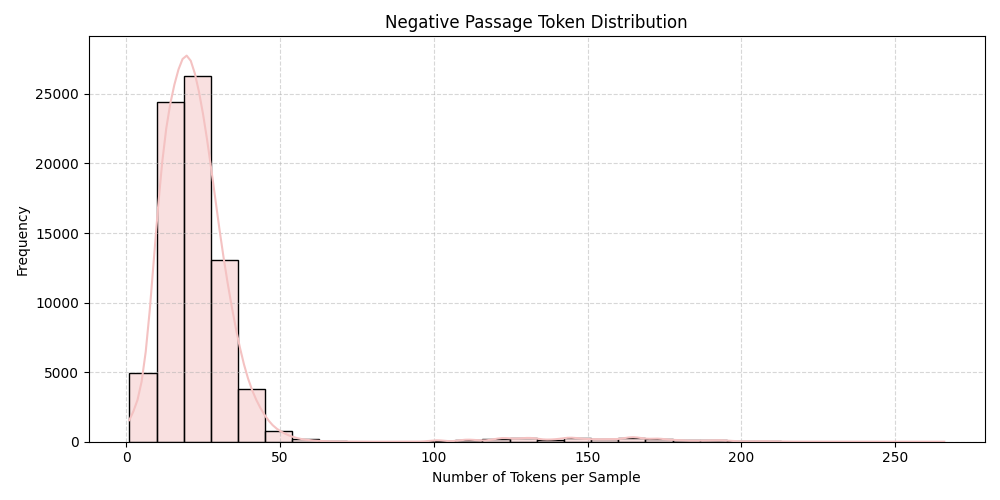}
        \caption{Negative tokens}
        \label{fig:neg_token_dist}
    \end{subfigure}
    \caption{Token count distributions across query, positive, and negative responses in the fine-tuning dataset.}
    \label{fig:token_distributions}
\end{figure*}

Additionally, we visualize dataset structure via three distribution plots:
(1) Figure~\ref{fig:topic_distribution} illustrates the estimated question distribution across key telecom topics,
(2) Figure~\ref{fig:query_token_dist} displays the token distribution in queries, and
(3) Figures~\ref{fig:pos_token_dist} and \ref{fig:neg_token_dist} show token count distributions for positive and negative responses respectively.
These plots provide insights into both the linguistic complexity and topical balance of the dataset.

\begin{table}[htbp]
\centering
\caption{Token statistics across dataset components.}
\label{tab:token_stats}
\resizebox{\columnwidth}{!}{%
\begin{tabular}{@{}lccc@{}}
\toprule
\textbf{Field}     & \textbf{Avg. Tokens} & \textbf{Min Tokens} & \textbf{Max Tokens} \\
\midrule
Query              & 11.26                & 8                   & 58                  \\
Positive Response   & 50.26                & 14                  & 557                  \\
Negative Response   & 25.00                & 11                   & 266                 \\
\midrule
\textbf{Vocabulary Size} & \multicolumn{3}{c}{50,586} \\
\bottomrule
\end{tabular}%
}
\end{table}

\begin{figure*}[t]
    \centering

    \begin{subfigure}[t]{0.49\textwidth}
        \centering
        \includegraphics[width=\textwidth]{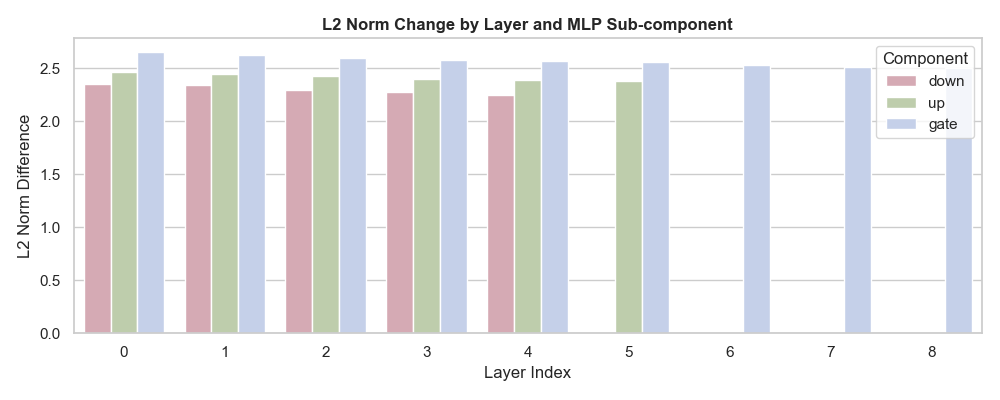}
        \caption{\textbf{Layer-wise Parameter Adaptation.} We plot the L$_2$‑norm differences $\Delta(W)=\|W_{\mathrm{fine}}-W_{\mathrm{base}}\|_2$ for each transformer layer (0–19), grouped by MLP sub‑component (gate, up‑proj, down‑proj). Gate projections exhibit the largest shifts across layers.}
        \label{fig:weight_change}
    \end{subfigure}
    \hfill
-
    \begin{subfigure}[t]{0.49\textwidth}
        \centering
        \includegraphics[width=\textwidth]{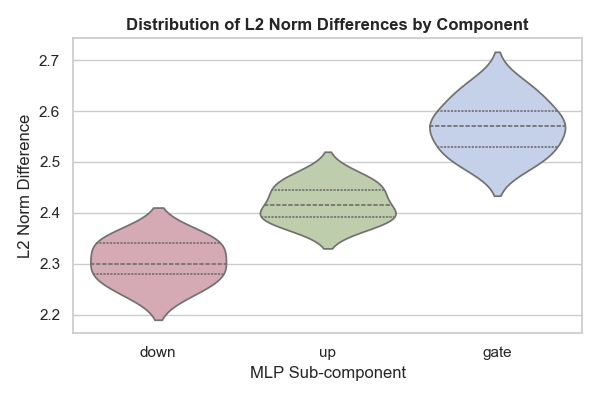}
        \caption{\textbf{Distribution of L$_2$ Norm Changes.} Violin plot visualizing the spread of parameter shifts across all MLP weight tensors. The wide base and sharp tails reflect high variability and the presence of deeply adapted subspaces.}
        \label{fig:violin_l2}
    \end{subfigure}

    \caption{\textbf{Visualization of Weight Adaptation.} Left: Per-layer changes highlight systematic adaptation in MLP sub-components. Right: Distributional view emphasizes the extent and variance of fine-tuning across model weights.}
    \label{fig:dual_l2_viz}
\end{figure*}
\section{Retrieval Dataset Details}

We constructed the \textbf{T-VEC Retrieval Dataset}, consisting of 1500 query-passage pairs, to evaluate telecom-specific knowledge retrieval capabilities. The dataset is derived from publicly available RFC documents and consists of natural-language queries paired with relevant technical passages. Each query is designed to retrieve a semantically appropriate excerpt, simulating realistic information-seeking scenarios in the telecom domain.

Query lengths average around 100 characters, while passage lengths vary more widely, with some exceeding 5,000 characters due to the inclusion of dense technical detail. Each passage includes provenance metadata such as the originating RFC ID, allowing traceability and potential reuse in downstream document-level tasks.

Each query is a natural-language question designed to retrieve a semantically relevant technical excerpt from an RFC document. Passages range in length from 80 to over 5,000 characters and include provenance metadata such as RFC source ID.

\section{Query Generation Prompt for Telecom Retrieval}
\label{query_generation_prompt}

To construct our domain-specific retrieval dataset, we formulated a prompt tailored to elicit high-quality, information-seeking queries grounded in telecommunications literature. Each prompt instance provides the model with a snippet from an RFC or related technical document and instructs it to generate a semantically relevant question that would ideally retrieve the given passage.

\paragraph{Prompt Template.}
\begin{quote}
You are a telecommunications expert assisting in the development of a technical search engine. Given a snippet from an RFC or vendor document, generate a question that would retrieve this snippet as a top-ranked result.

\textbf{Guidelines:}
\begin{itemize}
    \item The question must be answerable from the content of the passage, though it need not cover the entire snippet.
    \item Prefer domain-relevant, technical terminology, and use paraphrasing where possible instead of copying text verbatim.
    \item Avoid overly broad or overly specific questions. Keep the focus on key technical concepts present in the passage.
    \item Limit the query to a maximum of 20 words.
\end{itemize}

\textbf{Output Formatting:} Return only the query on a single line with no quotation marks, metadata, or explanation.
\end{quote}

\section{Evaluation on Standard Benchmarks}
\label{sec:evaluation_on_standard_benchmarks}

Although standard evaluation suites such as MTEB~\cite{muennighoff2022mteb} do not include telecom-specific corpora, we evaluate T-VEC and several baselines on a range of general-purpose semantic tasks. These include classic Semantic Textual Similarity (STS) datasets (STS12–STS16, STS-Benchmark), Natural Language Inference (NLI) via AllNLI, and additional retrieval-oriented tasks from MTEB such as ArguAna and SciDocsRR.

\paragraph{Semantic Textual Similarity (STS).}
We report Spearman correlation ($\times 100$) across eight standard STS tasks in Table~\ref{tab:sts_results}. Spearman’s rank correlation coefficient $\rho$ measures the strength of the monotonic relationship between two ranked variables. 
As a non-parametric metric, Spearman correlation is robust to non-linear relationships and is widely adopted for evaluating semantic similarity\cite{cer-etal-2017-semeval}. This makes it especially suitable for STS tasks where the goal is to capture relative semantic closeness rather than absolute distance. T-VEC consistently matches or outperforms its base model (GTE), achieving the highest average score across all datasets.

\renewcommand{\arraystretch}{0.90}  
\begin{table}[h]
  \centering
  \caption{Estimated average performance of various embedding models on the MTEB benchmark suite. Higher scores indicate better average task performance.}
  \label{tab:mteb_scores}
  \begin{tabular}{@{}lc@{}}
    \toprule
    \textbf{Model Name}                   & \textbf{Avg. Score} \\
    \midrule
    \textbf{T‑VEC}                        & \textbf{0.825}      \\
    bge‑base‑en‑v1.5                       & 0.815               \\
    gte‑base                             & 0.805               \\
    gte‑Qwen2‑1.5B‑instruct              & 0.795               \\
    instructor‑xl                        & 0.785               \\
    e5‑base‑v2                           & 0.780               \\
    jina‑embeddings‑v2‑base‑en           & 0.775               \\
    all‑mpnet‑base‑v2                     & 0.770               \\
    multilingual‑e5‑base                 & 0.765               \\
    all‑MiniLM‑L6‑v2                      & 0.760               \\
    \bottomrule
  \end{tabular}
\end{table}
\begin{table}[h]
  \centering
  \setlength{\tabcolsep}{3pt}
  \caption{Performance on the \texttt{sentence-transformers/all-nli} triplet evaluation. Higher scores indicate better general‑domain NLI capabilities.}
  \label{tab:allnli}
  \begin{adjustbox}{width=0.9\linewidth,center}
    \begin{tabular}{@{}lc@{}}
      \toprule
      \textbf{Model Name}                  & \textbf{Avg. Triplet Score} \\
      \midrule
      all‑mpnet‑base‑v2                     & 0.9620 \\
      bge‑base‑en‑v1.5                       & 0.9610 \\
      jina‑embeddings‑v2‑base‑en           & 0.9590 \\
      gte‑base                             & 0.9470 \\
      all‑MiniLM‑L6‑v2                      & 0.9380 \\
      e5‑base‑v2                           & 0.9230 \\
      instructor‑xl                        & 0.9220 \\
      multilingual‑e5‑base                 & 0.9210 \\
      gte‑Qwen2‑1.5B‑instruct              & 0.8660 \\
      \textbf{T‑VEC}                       & \textbf{0.6150} \\
      \bottomrule
    \end{tabular}
  \end{adjustbox}
\end{table}
\begin{table*}[t]
    \centering
    \footnotesize
    \caption{STS performance comparison of T-VEC against baseline and recent embedding models. Scores are Spearman correlation ($\times$100).}
    \label{tab:sts_results}
    \resizebox{\textwidth}{!}{%
    \begin{tabular}{l c c c c c c c c c c}
        \toprule
        \textbf{Task} & \textbf{T-VEC} & Qwen2 & BGE & MPNet & GTE & E5 & Instr. & Jina & MiniLM & MultiE5 \\
        \midrule
        ArguAna & 61.15 & 62.34 & \textbf{63.62} & 46.52 & 57.15 & 51.60 & 54.88 & 44.15 & 50.17 & 47.83 \\
        SciDocsRR & 83.97 & 81.56 & 87.49 & \textbf{88.65} & 87.08 & 82.83 & 79.54 & 83.11 & 87.12 & 80.39 \\
        STS12 & \textbf{80.32} & 72.81 & 78.03 & 72.63 & 75.71 & 73.49 & 74.08 & 74.28 & 72.37 & 77.93 \\
        STS13 & \textbf{88.22} & 84.70 & 84.18 & 83.48 & 85.73 & 83.00 & 85.05 & 84.18 & 80.60 & 76.89 \\
        STS14 & \textbf{82.75} & 78.80 & 82.27 & 78.00 & 81.51 & 80.45 & 80.32 & 78.81 & 75.59 & 77.53 \\
        STS15 & 88.26 & 87.45 & 87.96 & 85.66 & \textbf{88.81} & 88.18 & 88.36 & 87.55 & 85.39 & 88.37 \\
        STS16 & 84.78 & 84.94 & \textbf{85.47} & 80.03 & 83.82 & 83.66 & 83.78 & 85.35 & 78.99 & 82.70 \\
        STS-B & \textbf{88.05} & 85.38 & 86.42 & 83.42 & 85.74 & 85.48 & 83.05 & 84.84 & 82.03 & 84.20 \\
        \midrule
        \textbf{Average} & \textbf{82.19} & 79.75 & 81.93 & 77.30 & 80.69 & 78.59 & 78.63 & 77.78 & 76.53 & 76.98 \\
        \bottomrule
    \end{tabular}
    }
\end{table*}

\paragraph{Natural Language Inference.}

To assess generalization beyond the target domain, we evaluated all models on the AllNLI benchmark using the \texttt{sentence-transformers/all-nli} dataset. As shown in Table~\ref{tab:allnli}, T-VEC underperforms models trained specifically for general-domain NLI, reflecting a typical trade-off: domain specialization improves in-domain performance at the potential cost of generalization~\cite{gururangan-etal-2020-dont, howard-ruder-2018-universal, lee2020biobert}. Prior work has shown that aggressive fine-tuning on domain-specific corpora can lead to feature overfitting, where models excel in narrow contexts but degrade on out-of-distribution (OOD) or general-domain tasks.

\paragraph{Overall MTEB Performance.}

To provide a broader view of overall embedding quality, Table~\ref{tab:mteb_scores} reports the estimated average performance of each model across the MTEB benchmark suite. T-VEC ranks highest on this aggregate metric, suggesting strong generalization despite its domain-specific training. Table \ref{tab:mteb_scores} reports averages computed over a selected subset of MTEB v1 tasks (e.g., Arguana, SciDocs), rather than the full leaderboard set. The high average is largely driven by strong gains on retrieval-oriented tasks (e.g., SciDocsRR, STS13–15). Small variations also arise from the probabilistic nature of evaluation (e.g., stochasticity in negative mining and training).

\end{document}